# Knowledge Structures and Evidential Reasoning in Decision Analysis

*Gerald S. Liu\**

Computer Aided Systems Laboratory
Schlumberger Palo Alto Research
3340 Hillview Avenue
Palo Alto, CA 94304

Decision Making.

*ABSTRACT*

In the context of sophisticated decision making, the distinct roles played by decision factors are characterized in terms of their behavior in affecting the final decision. The evidence that partially matches a factor is considered. Then effective computation rules are presented to aggregate the evidence in order. It is argued that a deeper level of causality can be expressed, and that the cognitive structure may be better preserved.

## 1. Introduction

Multi-attribute utility theory ([van Neumann 1947], [Fishburn 1970]) has long been an important tool for classical decision makings. Together with the rational *homo economicus* model, it has received numerous criticisms (e.g. [Simon 1947], [Edwards 1954], etc). Indeed, in the reality of subjective decision making, individuals as the decision makers can seldom afford to select among several alternatives (as in [Tversky 1972], for instance). Instead, a primarily bivalent conclusion is often needed for the particular case under consideration. Besides, very often the rich body of causal relations between the factors and the decision is lost in the simple, conventional model of factors weighting (e.g. [Fishburn 1967], [Yager 1977]). To address these issues, in this paper, we shall de-emphasize the aspect of optimal choice in the decision, and elaborate the sophisticated process of case analysis and evaluations. To this end, the earlier studies of *confirmation* (e.g. [Carnap 1950]) become more relevant in spirit (but not to the extent of their appeal to probabilities, for reasons presented in the next section.)

Since we concentrate on the decision making as a process of subjective judgement, the burgeoning field of expert systems naturally provides many useful insights. In particular, early systems like MYCIN [Shortliffe 1976] (which will be frequently referred to in the sequel,) and PROSPECTOR [Duda 1979] have explored various relationships between the evidence and the hypothesis. Lately, the theoretical works by [Dempster 1967] and [Shafer 1976] that extend probability theories have received wide attention for their capability of capturing incomplete and uncertain knowledge. (See section 3.)

Recently, from a different angle, [Cohen 1983] has argued that the straightforward numerical representation cannot discriminate kinds of causative situations. Therefore a decision system that remains entirely numerical cannot catch the deep meaning of factor causality. In part, this argument coincides with our previous criticism against the conventional factor-weighting mechanism. However, between the two extremes of plain numerical calculation and pure symbolic manipulation, (which Cohen's work rested on,) we envisage a great deal of latitude for improvement: On the one hand, we acknowledge that different pieces of evidence do exhibit distinct natures, therefore recklessly massaging them into a single number without regard to this heterogeneity is oversimplistic. On the other hand, we also regard aggregating different factors as an indispensable capability for an intelligent system to reach conclusions.

It is from this stand that we shall present this paper.

## 2. Factors and Evidence

During the process of analysis and judgement, we recognize the existence of a subjective tendency in granting the decision. (At this moment we content ourselves with the intuitive meaning.) In addition, the decision tendency will change as the situation presents itself by means of different factors. As mentioned in the introduction, the causal relations are manyfold between the fac-

---

tors and the subjective tendency to granting the decision *before* and *after* acknowledging a factor. In order to describe these relations, we introduce the following notations:

> H: that the decision maker will make a positive decision on the candidate under consideration;

> *F is V*: a factor F takes the value V; (This factor-value association generalizes the bivalent judgement in which a piece of fact $F=V_0$ would be either true or false. The reason for this generalization is that usually it is not only the factor existing or not that counts, but also *how* the factor exists. As a matter of fact, different factor values can change the way it influences the decision; in other words, they may change the role this factor is playing. Similar concepts on this many-valued variable were formalized in a variable-valued logic calculus $VL_1$ for an expert system on soybean disease diagnosis. [Michalski 1980])

> $Bel(H|E)$: the *prior degree of belief* that H should be true, given some *other* knowledge E irrelevant of F. In other words, $Bel(H|E)$ expresses the subjective *tendency* of the decision maker to grant a positive decision; (It should be noted that we are *not* following the Bayesian theory, therefore $Bel(H|E)$ doesn't obey the usual Bayesian *rule of conditioning* for probabilities: $\frac{Bel(H \cap E)}{Bel(E)}$. A better interpretation in light of Dempster and Shafer's works will be presented in the sequel.)

> $Bel(H|E, F\ is\ V)$: the *posterior degree of belief* that H should be true, given E and the *incremental* knowledge that F is V;

It was pointed out earlier that $Bel(H|E)$ should not be equated to the conditional probability $Prob(H|E)$ - or just the prior probability $Prob(H)$ in case of E being trivially true. The major distinction between $Bel(H|E)$ and $Prob(H|E)$ - assuming the latter does cognitively make sense for the moment - is that we require $Bel(H|E)+Bel(\overline{H}|E)\leq 1$ in general. This being the case, one possible interpretation may be to identify $Bel(H|E)$ with the *belief function* in Dempster-Shafer theory. Indeed, some computational formulae in this paper bear a close relation to their counterparts in Dempster-Shafer theory, which will be examined in details in section 3.

### 2.1 The Role System of Factors

In this section, all the decision factors will first be categorized into different *elementary roles* (including *supportive, adverse, sufficient, contrary,* and *necessary*) or their combinations when meaningful. These different roles together constitute a role system that sorts out the corpus of human knowledge so that factors of distinct nature can be treated differently. For each of the five elementary roles, we additionally specify an *intensity*, indicating how strongly the role may affect the final decision, and two *margins*, signifying the valid range over which the factor would play this role. In addition, multiple roles can be played by the same factor, in which case the composite effect is amenable to a *rule of superposition* on the elementary roles.

### (1) Supportive Factor

A supportive factor doesn't have to be present in order to grant a positive decision. In fact, the absence of a purely supportive factor should not count against the decision at all. But if the supportive factor does exist, then it should contribute to confirming the decision with a *degree of support* (the intensity).

In analogy to the *Measure of increased Belief (MB)* in MYCIN, we have tried to capture the concept of degree of support (SUPP) by the following equation, with the understanding that $Bel(H|E, F\ is\ V)$ is larger than $Bel(H|E)$ under the circumstances:

$$SUPP = \frac{Bel(H|E, F\ is\ V) - Bel(H|E)}{1 - Bel(H|E)} \quad (1)$$

It should be noted that the degree of support is an intuitive concept. When estimated by a decision maker, it reflects in part the cognitive structure of this person. It therefore remains characteristic of him only. Another decision maker may very well provide a different value as the degree of support by the same factor. Contrary to the possibility that this will introduce an undesirable discrepancy from person to person, the variation accounts for the individual subjectivity that is intrinsic of a personal decision making.

With an estimated degree of support SUPP from the decision maker, and with two *margins* $V_L$ and $V_H$ (that form the valid value range $|V_L, V_H|$), we then can completely describe the behavior of a supportive factor F by the following pair of equations:

$$Bel(H|E, F\ is\ V_H) = Bel(H|E) + SUPP \times (1 - Bel(H|E)) \quad (2)$$

$$Bel(H|E, F\ is\ V_L) = Bel(H|E)$$

The distinction between our use of the degree of support SUPP and MYCIN's measure of belief MB is now made clear: in MYCIN, although MB was *defined* in terms of prior and posterior probabilities, the system never really attempted to calculate them. Instead, ad hoc formulae were invented that only involved MB's themselves. As a result, it doesn't really make a difference how MB was defined. In contrast, the primary concern in this paper is to update the posterior degree of belief using estimated SUPP's and other kinds of role intensities. (It should be noted that two later studies in [Ishikuza 1981] and [Adams 1984] did try to derive MB's based on probabilities and Bayes' theorem. However, the former concluded with slightly different combination functions.)

### (2) Adverse Factor

An adverse factor is a factor which counts against the decision, not to the full, but with a *degree of adversity* (i.e. the intensity). It should be pointed out that the absence of an adverse factor simply means that there is no disconfirmatory effect due to this factor. This is not to be confused with the presence of a supportive factor, in which case the degree of belief in the decision would actually be increased.

Similar to the way SUPP was defined, the degree of adversity ADV by factor F can be expressed by the following equation a la MYCIN, provided that $Bel(H|E, F$ is $V)$ is known to be smaller than $Bel(H|E)$:

$$ADV = \frac{Bel(H|E) - Bel(H|E, F \text{ is } V)}{Bel(H|E)} \quad (3)$$

An interesting observation is in order at this juncture. Although the definitions of SUPP and ADV bear a close resemblance to the MYCIN definitions of MB and MD (Measure of increased Belief) respectively, many properties therein do not hold any more. For instance, if we read $SUPP(H|F$ is $V)$ as the degree of support for H by F being V, and $ADV(H|F$ is $V)$ as the degree of adversity against H by F being V, then it is *not* true that $SUPP(H|F$ is $V) = ADV(\overline{H}|F$ is $V)$, while in MYCIN we did have $MB(H,E) = MD(\overline{H},E)$ for any evidence E. The primary reason for this difference comes from our requirement that $Bel(H|E) + Bel(\overline{H}|E) \leq 1$ as previously stated. Less mathematically neat than $Prob(H|E) + Prob(\overline{H}|E) = 1$ as it may appear at first, the new formalism is free from a very undesirable consequence in MYCIN: $MD(H|E) > 0$ implies $MB(H|\overline{E}) > 0$; in other words, the absence of an adverse factor in MYCIN would amount to the presence of a supportive factor! [Yen 1985]

In regard to the man-machine interaction, the value of ADV, just like the value of SUPP, must also be solicited from the decision maker. With its degree of adversity determined, an adverse factor F can then be completely described in terms of its bearings on the posterior belief as follows:

$$Bel(H|E, F \text{ is } V_H) = Bel(H|E) \times (1 - ADV) \quad (4)$$
$$Bel(H|E, F \text{ is } H_L) = Bel(H|E)$$

### (3) Sufficient Factor

A sufficient factor is such that its confirmation *alone* suffices to merit an affirmative decision, but no confirming effect will be observed if a pure sufficient factor doesn't exist. In practice, the *degree of sufficiency* (SUFF) may be less than perfect. Therefore we may use the following pair of equations to characterize a sufficient factor:

$$Bel(H|E, F \text{ is } V_H) = max(Bel(H|E), SUFF) \quad (5)$$
$$Bel(H|E, F \text{ is } V_L) = Bel(H|E)$$

Care must be taken here to distinguish a *strongly* supportive factor from its sufficient counterpart that is somewhat weak. In essence a supportive factor *incrementally* adds to the total degree of belief, while a sufficient factor *ensures* the belief to quite a considerable degree. The reason for this distinction is that a supportive factor, unlike its sufficient counterpart entailing by itself all the information that is needed, usually is intended to *corroborate* with other supportive factors. This need and capability for corroboration remains characteristic of all the supportive factors, even if a particular supportive factor is so strong that it corroborates to yield a very high degree of belief as a single sufficient factor would. Since rarely is there no need for corroborative considerations, the supportive role is generally preferred.

### (4) Necessary Factor

The factor has to be present in order to make a positive decision. If the factor is absent, however, the belief in the decision will be reduced to nil. On the other hand, the confirmation of a pure necessary factor does not contribute to a larger degree of belief at all.

In practice, we shall describe the behavior of a necessary factor by means of its *degree of necessity* NEC defined as: $NEC = 1 - Bel(H|E, F \text{ is } V_L)$. Note that $Bel(H|E, F \text{ is } V_L)$ generalizes the classical notation $Prob(H|E, \overline{F})$ and should be close to 0. Now that an estimated degree of necessity is given by the decision maker, we are then in a position to describe a necessary factor F as follows: (Remember that $F$ is $V_H$ generalizes $F$ is *present*, and $F$ is $V_L$ generalizes $F$ is *absent* ($\overline{F}$).)

$$Bel(H|E, F \text{ is } V_H) = Bel(H|E) \quad (6)$$
$$Bel(H|E, F \text{ is } V_L) = min(Bel(H|E), 1 - NEC)$$

### (5) Contrary Factor

If a contrary factor is present, the degree of belief in the positive decision will be excluded. However, the absence of such a factor doesn't lead to an increased belief. Thus, a contrary factor in a sense is the negation of a necessary factor.

Similar to the degree of necessity, we can define a *degree of contraries* CONTR as: $CONTR = 1 - Bel(H | E, F \text{ is } V_H)$, which will be again solicited from the decision maker. Then the contrary factor will act up to the following pair of equations:

$$Bel(H | E, F \text{ is } V_H) = min(Bel(H | E), 1 - CONTR) \qquad (7)$$

$$Bel(H | E, F \text{ is } V_L) = Bel(H | E)$$

Although the same factor can play multiple roles, it should be noted that an intuitive rule of superposition precludes any contradictory combination. For example, it is not allowed to have the same factor be both supportive and adverse. But a supportive role can certainly be combined with a necessary role.

### 2.2 Combination of Factors

In the previous section, we have studied the individual behaviors of different kinds of roles for a factor; we have also remarked that multiple roles can be played by the same factor within the realm of logical consistency. However, the issue of the joint behavior by multiple factors was not addressed. Since very rarely would there be a single-factor decision situation - even if the factor consists of compound statements, the study of the issue in this section thus becomes imperative.

**(1) Multiple Supportive Factors**

Suppose we have two supportive factors $F_1$ and $F_2$ respectively defined as follows:

$$Bel(H | E, F_1 \text{ is } V_{H_1}) = Bel(H | E) + S_1 \times (1 - Bel(H | E)) \qquad (8)$$

$$Bel(H | E, F_1 \text{ is } V_{L_1}) = Bel(H | E)$$

$$Bel(H | E, F_2 \text{ is } V_{H_2}) = Bel(H | E) + S_2 \times (1 - Bel(H | E)) \qquad (9)$$

$$Bel(H | E, F_2 \text{ is } V_{L_2}) = Bel(H | E)$$

Our interest is to calculate the *joint posterior degree of belief*, formally $Bel(H | E, F_1 \text{ is } V_{H_1}, F_2 \text{ is } V_{H_2})$. If we recall the meaning of $Bel(H | E)$, this is easily answered by substituting $Bel(H | E, F_1 \text{ is } V_{H_1})$ for $Bel(H | E)$ in equation (9). After rearrangement, we obtain the following pair of equations for the joint belief:

$$Bel(H | E, F_1 \text{ is } V_{H_1}, F_2 \text{ is } V_{H_2}) \qquad (10)$$

$$= Bel(H | E) + (S_1 + S_2 - S_1 \times S_2) \times (1 - Bel(H | E))$$

$$Bel(H | E, F_1 \text{ is } V_{L_1}, F_2 \text{ is } V_{L_2}) = Bel(H | E)$$

It is interesting to observe that the *joint degree of support* SUPP obeys the following relation with the two constituent degrees of support $S_1$ and $S_2$:

$$SUPP = S_1 \odot S_2 = S_1 + S_2 - S_1 \times S_2 \qquad (11)$$

Not surprisingly, this combination method bears exactly the same functional form as the combined measures of belief in MYCIN. However, because our derivation is based on the increased belief resulting from a factor valuation, we are able to perform *partial* matchings against available evidence, as will be examined in the next section. Finally, we shall just briefly remark that the sequence in combining multiple supportive factors does not affect the final posterior belief, due to the fact that the combination formula is both commutative and associative.

**(2) Multiple Adverse Factors**

Similar to the case with multiple supportive factors, suppose we have two adverse factors $F_1$ and $F_2$ defined by:

$$Bel(H | E, F_1 \text{ is } V_{H_1}) = Bel(H | E) \times (1 - A_1) \qquad (12)$$

$$Bel(H | E, F_1 \text{ is } V_{L_1}) = Bel(H | E)$$

$$Bel(H | E, F_2 \text{ is } V_{H_2}) = Bel(H | E) \times (1 - A_2) \qquad (13)$$

$$Bel(H | E, F_2 \text{ is } V_{L_2}) = Bel(H | E)$$

The question here is again to ask the value of the joint posterior degree of belief $Bel(H | E, F_1 \text{ is } V_{H_1}, F_2 \text{ is } V_{H_2})$. By substituting $Bel(H | E, F_1 \text{ is } V_{H_1})$ in equation (12) for $Bel(H|E)$ in (13), we easily derive the following relations:

$$Bel(H | E, F_1 \text{ is } V_{H_1}, F_2 \text{ is } V_{H_2}) \qquad (14)$$

$$= Bel(H | E) \times (1 - A_1) \times (1 - A_2)$$

$$Bel(H | E, F_1 \text{ is } V_{L_1}, F_2 \text{ is } V_{L_2}) = Bel(H | E)$$

Just like the joint degree of support in the previous case, from (14) we can also derive a *joint degree of adversity* ADV in terms of $A_1$ and $A_2$ such that $Bel(H | E, F_1 \text{ is } V_{H_1}, F_2 \text{ is } V_{H_2}) = Bel(H | E) \times (1 - ADV)$: (Therefore although our SUPP and ADV lack the same symmetry with respect to H as in MB and MD, we still enjoy an identical combination method between SUPP's and ADV's respectively, hence the combination sequence for adverse factors is also commutative and associative.)

$$ADV = A_1 \bigcirc A_2 = A_1 + A_2 - A_1 \times A_2 \quad (15)$$

**(3) Supportive Factors in Conjunction with Adverse Factors**

Consider the scenario in which both a supportive factor and an adverse factor come into play. In other words, suppose we have equations (8) and (13) both being effective for the decision maker, with factor $F_1$ being supportive and factor $F_2$ adverse, then what is the joint degree of belief $Bel(H|E, F_1 \text{ is } V_{H_1}, F_2 \text{ is } V_{H_2})$ in this case?

At this juncture it seems that we are entitled to two alternatives for deriving the joint belief: in the first approach we substitute $Bel(H|E, F_1 \text{ is } V_{H_1})$ for $Bel(H|E)$ in equation (13), in which case the supportive factor will be *discounted* by the adverse factor (Remember that equation (13) relates to the adverse factor $F_2$). In the second approach, however, we substitute $Bel(H|E, F_2 \text{ is } V_{H_2})$ for $Bel(H|E)$ in equation (8), in which case we shall have the adverse factor *lessened* by its supportive counterpart. Unfortunately, the two methods will yield different results, and it can be easily shown that the first method *always* leads to a more conservative (smaller) degree of belief. So, which is the one to choose?

It turns out that the first method - supportive factors discounted by adverse ones - is better for the following two reasons. First, it always incorporates both the supportive and the adverse effects in the calculation, regardless of the initial value of $Bel(H|E)$. In contrast, the second method will suppress the adverse effect in case of the initial $Bel(H|E)$ being 0, this is because we substitute $Bel(H|E, F_2 \text{ is } V_{H_2}) = Bel(H|E) \times (1-A_2) = 0$ for $Bel(H|E)$ in equation (8). Second, by using Dempster-Shafer theory with appropriate interpretations it is possible to derive the same result as in the first method. (Our attempt to relate to that theory is covered in section 4.)

As a result, we shall adopt the first method to derive the joint degree of belief as follows:

$$Bel(H|E, F_1 \text{ is } V_{H_1}, F_2 \text{ is } V_{H_2}) \quad (16)$$
$$= [Bel(H|E) + S_1 \times (1 - Bel(H|E))] \times (1 - A_2)$$
$$Bel(H|E, F_1 \text{ is } V_{L_1}, F_2 \text{ is } V_{L_2}) = Bel(H|E)$$

Contrary to the previous cases with multiple supportive or adverse factors, in which the joint degree of support/adversity can be expressed in terms of respective components as in (11) and (15), there is no simple counterpart here - neither a joint degree of support or a joint degree of adversity is composable without involving the $Bel(H|E)$ terms. This observation confirms a common criticism that the MYCIN formula for Certainty Factors $CF = MB - MD$ when combining confirmatory and disconfirmatory evidence is truly an ad hoc device. In addition, the same argument also applies to the EMYCIN revision [van Melle 1980], which is $CF = \frac{MB - MD}{1 - min(MB, MD)}$.

**(4) Multiple Sufficient, Necessary, or Contrary Factors**

Because of the *nonaggregative* nature, as mentioned earlier, in these three kinds of factors, the combination of their instances is therefore different from that of supportive or adverse factors. To start with, let us consider the situation where there are two sufficient factors $F_1$ and $F_2$, with respective degrees of sufficiency $SUFF_1$ and $SUFF_2$. Then the joint degree of belief may be established by the factor with a larger degree of sufficiency. Stated mathematically, with the notation $SUFF_{MAX} = max(SUFF_1, SUFF_2)$, we shall have

$$Bel(H|E, F_1 \text{ is } V_{H_1}, F_2 \text{ is } V_{H_2}) = max(Bel(H|E), SUFF_{MAX}) \quad (17)$$
$$Bel(H|E, F_1 \text{ is } V_{L_1}, F_2 \text{ is } V_{L_2}) = Bel(H|E)$$

Similarly, the joint degree of belief for two necessary factors may be defined as

$$Bel(H|E, F_1 \text{ is } V_{H_1}, F_2 \text{ is } V_{H_2}) = Bel(H|E) \quad (18)$$
$$Bel(H|E, F_1 \text{ is } V_{L_1}, F_2 \text{ is } V_{L_2}) = min(Bel(H|E), 1 - NEC_{MAX})$$

where $NEC_{MAX} = max(NEC_1, NEC_2)$; by the same token, two contrary factors should act upon the joint belief according to the following pair of equations:

$$Bel(H|E, F_1 \text{ is } V_{H_1}, F_2 \text{ is } V_{H_2}) \quad (19)$$
$$= min(Bel(H|E), 1 - CONTR_{MAX})$$
$$Bel(H|E, F_1 \text{ is } V_{L_1}, F_2 \text{ is } V_{L_2}) = Bel(H|E)$$

**(5) The Overall Combination**

We have studied several combinations between different factors, but we still have to determine the overall combining sequence that includes the entirety of decision factors. From previous discussions, it is natural to propose a sequence such that all supportive factors will be combined first, followed by successive discounts from the adverse factors. Finally the total degree of belief is either guaranteed by sufficient factors, or nullified by contrary factors, or else filtered by the necessary ones.

**2.3 Partial Matching by Evidence**

277

In previous sections we have presented a role system for factors considered to be relevant by the decision maker. Although different roles assume different behaviors, it is common to all of them that the factor (as the role player) represents only a generic expectation *beforehand*. As the time comes when actually in the decision process, the judgement made on each factor really depends on the particular *evidence* obtained; whether done reliably or by virtue of guesses, the "evidence" will be used to match against the corresponding factor.

This factor/evidence dichotomy can also be viewed as patterns versus data (in the terminology of pattern matching), or decision variables versus their values (in the classical theory). At any rate, the point is that the matching may not be perfect, therefore the asserted *rule of behavior* (i.e. the paired equations) may not apply to the whole. Then it is really up to the decision maker to determine the effectiveness of the original rule. If the person performs a bivalent, all-or-none matching (which we shall call a *rigid judgement*), then all the evidence that falls short of expectation will amount to nothing, no matter how small the margin is. Alternatively, the decision maker can release the rigorous constraint and employs an *elastic judgement*, in which case the rule of behavior will be *partially* observed on imperfect evidence. In this paper, we assume that *all* the judgements are elastic. This is purely for the convenience of discussion.

Suppose a particular piece of evidence shows that $F$ is $V_E$ in the range of factor values $[V_L, V_H]$. As a result several cases may arise in light of the theory of measurement (e.g. [Krantz 1971]). The factor values may be in *nominal* or *ordinal* scale, in which case it would be difficult to reason the behavior of $F$ is $V_E$, short of equations for F at values other than $V_L$ or $V_H$. However, if the factor values are at least in *interval* scale, then it makes more sense to measure how closely the evidence is present to the full degree. This measurement, which will be referred to as the *evidential strength* $\eta$, may be defined by the following ratio:

$$0 \leq \eta = \frac{V_E - V_L}{V_H - V_L} \leq 1 \qquad (20)$$

(Therefore the definition of this evidential strength does generalize the rigid judgement to be more than a matter of absence/presence.) Now that we have computed $\eta$ from the evidence (which we shall write as $F$ is $V_E$), we are facilitated to reason the new rule of behavior for factor F under various circumstances. For instance, if F is a supportive factor the posterior belief Bel(H|E, F is V sub E ) can be estimated by a *linear interpolation* between the two equations in (2):

$$Bel(H|E, F \text{ is } V_E) = Bel(H|E) + SUPP \times \eta \times (1 - Bel(H|E)) \qquad (21)$$

It should be noted here that the linear interpolation has been chosen mainly for its simplicity; actually other methods are also applicable. In particular, if we replace $\eta$ by $\eta^n$ with a big n in the estimating equation (21), we are effectively *simulating* a rigid judgement. In fact this raise-to-powers method has been widely used in [Zadeh 1978a] and other related works to deal with various degrees of stringency in semantics.

In the case when F is an adverse factor evidenced with strength $\eta$, a linear interpolation between equations (4) would yield:

$$Bel(H|E, F \text{ is } V_E) = Bel(H|E) \times (1 - ADV \times \eta) \qquad (22)$$

When it comes to sufficient factors, a linear interpolation should be taken between Bel(H|E) and SUFF in the first equation of (5). The result then becomes

$$Bel(H|E, F \text{ is } V_E) \qquad (23)$$
$$= max(Bel(H|E), (1-\eta) \times Bel(H|E) + \eta \times SUFF)$$

Suppose F is a necessary factor, then a linear interpolation in equations (6) gives us the following estimation:

$$Bel(H|E, F \text{ is } V_E) \qquad (24)$$
$$= min(Bel(H|E), \eta \times Bel(H|E) + (1-\eta) \times (1 - NEC))$$

Finally, in the case of F being a contrary factor with strength $\eta$, we may obtain the following interpolated belief from (7):

$$Bel(H|E, F \text{ is } V_E) \qquad (25)$$
$$= min(Bel(H|E), (1-\eta) \times Bel(H|E) + \eta \times (1 - CONTR))$$

### 2.4 Aggregation of Evidence

In the previous section we have made a distinction between the factor and the evidence, accordingly the rules of the factor behavior were in essence discounted in proportion to the available evidential strengths. Note that the rule discount may be elastic to various degrees. This elasticity can be achieved by choosing the power term for interpolation. However, we shall confine ourselves to the linear case in this paper.

In more general settings, just as the decision maker needs to combine multiple factors specified, he often has to aggregate *pieces* of evidence collected. It is this issue that we now turn to.

Consider two pieces of evidence $E_1$: $F_1$ is $V_{E_1}$ and $E_2$: $F_2$ is $V_{E_2}$, with corresponding evidential strengths $\eta$ and $\varsigma$ defined by: $\eta = \frac{V_{E_1} - V_{L_1}}{V_{H_1} - V_{L_1}}$ and $\varsigma = \frac{V_{E_2} - V_{L_2}}{V_{H_2} - V_{L_2}}$. Our purpose is to estimate the *aggregated degree of belief* due to the two pieces of evidence, formally written as $Bel(H | E, F_1 \text{ is } V_{E_1}, F_2 \text{ is } V_{E_2})$, and we shall examine various cases in which $F_1$ and $F_2$ assumes various roles. To begin with, let us suppose they both are supportive factors, with respective degrees of support $S_1$ and $S_2$, then an inspection into equations (10), (11), and (21) will lead to the following equation:

$$Bel(H | E, F_1 \text{ is } V_{E_1}, F_2 \text{ is } V_{E_2}) \qquad (26)$$
$$= Bel(H | E) + SUPP \times (1 - Bel(H | E))$$

with the *aggregated degree of support* $SUPP$ equal to $S_1\eta + S_2\varsigma - S_1\eta \times S_2\varsigma$.

As a possible generalization, we can redefine this $SUPP$ to be $S_1 + S_2 - S_1 S_2 \times (\eta * \varsigma)$, where the two-place function $\eta * \varsigma$ is such that (1) $\eta * \varsigma = \varsigma * \eta$, (2) $(\eta * \varsigma) * \xi = \eta * (\varsigma * \xi)$, (3) if $\eta \leq \varsigma$ and $\mu \leq \nu$, then $\eta * \mu \leq \varsigma * \nu$, and (4) $1 * \eta = \eta$. Then the question of appropriate function forms for $\eta * \varsigma$ arises. It turns out that such $\eta * \varsigma$ has been widely studied in areas related to the fuzzy set theory, and is formally named a *triangular form*. The class of these triangular forms can take a variety of actual function forms, most notably $max(0, \eta + \varsigma - 1)$, $\eta \times \varsigma$, and $min(\eta, \varsigma)$, in an increasing order [Prade 1985]. Other *parametric* function forms of the class can be found in, e.g. [Dombi 1982]. For the purpose of illustration here, we shall refer to the one by Hamacher: $\eta * \varsigma = \frac{\eta \times \varsigma}{\lambda + (1-\lambda) \times (\eta + \varsigma - \eta \times \varsigma)}$ ($0 \leq \lambda \leq 1$). Notice that the $\eta * \varsigma$ above reduces to a simple product $\eta \times \varsigma$, which is its maximum for all possible $\lambda$'s, in the special case of $\lambda = 1$. Therefore by carefully adjusting the value of $\lambda$, we can control the product term $S_1 S_2 \times (\eta * \varsigma)$ in the aggregated degree of support $SUPP$, which really accounts for the *interaction* between the two pieces of evidence. It is for this reason that we regard $\lambda$ as a correlation indicator.

Similar result may be obtained when factors $F_1$ and $F_2$ are both adverse, since as observed earlier from equation (14), the joint degree of adversity employs the same combination formula as the joint degree of support. More specifically, with aid of equations (14) and (22), we shall have

$$Bel(H | E, F_1 \text{ is } V_{E_1}, F_2 \text{ is } V_{E_2}) = Bel(H | E) \times (1 - ADV) \qquad (27)$$

where $ADV$ is defined as $A_1\eta + A_2\varsigma - A_1 A_2 \times (\eta * \varsigma)$, with the understanding that $\eta * \varsigma$ generalizes the straightforward product term $\eta \times \varsigma$.

Now suppose factor $F_1$ is supportive, and $F_2$ is adverse. To compute the aggregated posterior belief $Bel(H | E, F_1 \text{ is } V_{E_1}, F_2 \text{ is } V_{E_2})$, we simply dissolve equations (21) and (22) into (16), then we shall obtain the following relation:

$$Bel(H | E, F_1 \text{ is } V_{E_1}, F_2 \text{ is } V_{E_2}) \qquad (28)$$
$$= [Bel(H | E) + S_1\eta \times (1 - Bel(H | E)] \times (1 - A_2\varsigma)$$

When it comes to two sufficient factors, with their nature being nonaggregative, the "aggregated" belief is then established by simply taking the maximum of the two interpolations. Stated symbolically, we have a relation very similar to (17):

$$Bel(H | E, F_1 \text{ is } V_{E_1}, F_2 \text{ is } V_{E_2}) = max(Bel(H | E), SUFF_{MAX}) \quad (29)$$

The only change is that $SUFF_{MAX}$ now becomes $max( (1-\eta) \times Bel(H | E) + \eta \times SUFF_1, (1-\varsigma) \times Bel(H | E) + \varsigma \times SUFF_2 )$.

By the same token, if both factors $F_1$ and $F_2$ are necessary, we shall have the counterpart of (18):

$$Bel(H | E, F_1 \text{ is } V_{E_1}, F_2 \text{ is } V_{E_2}) \qquad (30)$$
$$= min(Bel(H | E), 1 - NEC_{MAX})$$

with $NEC_{MAX} = 1 - min( \eta \times Bel(H | E) + (1-\eta) \times (1 - NEC_1), \varsigma \times Bel(H | E) + (1-\varsigma) \times (1 - NEC_2) )$. To better understand the meaning of the *aggregated degree of necessity* $NEC_{MAX}$ here, it is illuminating to study two special cases: (1) $\eta = \varsigma = 0$: this amounts to both factors being absent to the full. In this case $NEC_{MAX}$ is just the maximum of the two degrees of necessity. (2) $\eta = 1$ but $0 < \varsigma < 1$: this is the situation in which the decision maker will conclude that factor $F_1$ is completely fulfilled, but factor $F_2$ is somewhat absent. In other words, the decision maker can neglect the first factor and focuses on the partial evidential matching with the second. Indeed, simple algebraic operations would reduce equation (30) to the partial matching relation (24) for factor $F_2$. Very similar results can be derived for two contrary factors:

$$Bel(H | E, F_1 \text{ is } V_{E_1}, F_2 \text{ is } V_{E_2}) \qquad (31)$$
$$= min(Bel(H | E), 1 - CONTR_{MAX})$$

with $CONTR_{MAX} = 1 - min( (1-\eta) \times Bel(H | E) + \eta \times (1 - CONTR_1), (1-\varsigma) \times Bel(H | E) + \varsigma \times (1 - CONTR_2) )$.

As an ending remark in this section, the sequence in which we aggregate all pieces of evidence should be identical to the sequence in which all factors were combined.

### 3. A View from Dempster-Shafer Theory

In the beginning of section 2, we related $Bel(H|E)$ to the belief function in Dempster-Shafer theory. In this section, we shall examine the relation more comprehensively.

The major advantage of the Dempster-Shafer theory, as stated in [Gordon 1984], is "its ability to model the narrowing of the hypothesis set with the accumulation of evidence" in application to the diagnostic reasoning. Our purpose, being a bivalent decision making, we shall mainly benefit from a consequence of its generality. Specifically, the theory affords us to avoid the Bayesian restriction that the commitment of belief to a particular decision implies the commitment of the rest of belief to its negation; i.e. that $Prob(H|E)=1-Prob(\overline{H}|E)$.

To facilitate a partial interpretation of our model in view of the Dempster-Shafer theory, we shall first refer to some related concepts in that theory, then try to compare our model with them. It should be noted here that a familiarity with the theory itself is assumed, and that the concept mentions in the sequel are primarily meant to unify the notations before relating to our works.

Let us start with the *frame of discernment* $\Theta$, which, in our case, contains $H$ and $\overline{H}$ only. Then the *basic probability assignment* (bpa) $m$ will be such that $m(H)+m(\overline{H})+m(\Theta)=1$. Furthermore, over $H$ and $\overline{H}$ we shall have *belief functions* $B(H)$ and $B(\overline{H})$, which, as a special case in our situation, are identical to $m(H)$ and $m(\overline{H})$ respectively. (Although the whole issue of *belief* versus *plausibility* is of fundamental importance in the theory, it is not essential for our purpose of interpretation and therefore will not be addressed here.)

Let $B_1$ and $B_2$ denote two belief functions based on their bpa's $m_1$ and $m_2$ respectively. Dempster's rule of combination will then calculate the combined effect of $m_1$ and $m_2$, which is denoted by $m_1 \bigcirc m_2$. Based on this result, a combined belief function $B_1 \bigcirc B_2$ can be trivially obtained in our case. (Since the beliefs and the bpa's will take the same value over the singletons $H$ and $\overline{H}$.) In [Gordon 1984], several special cases in connection to MYCIN have been studied for the combined belief. For the purpose of reference in the sequel, we now rephrase two of them as follows:

Suppose $B_1(\overline{H})=0$ and $B_2(\overline{H})=0$, (In other words, both beliefs support $H$ only.) then the combined belief $B_1 \bigcirc B_2$ according to Dempster's rule will be such that

$$B_1 \bigcirc B_2(H) = B_1(H) + B_2(H) \times (1 - B_1(H)) \qquad (32)$$

On the other hand, if $B_1(\overline{H})=0$ and $B_2(H)=0$, which means the first belief supports $H$ while the second supports $\overline{H}$, then the combined belief $B_1 \bigcirc B_2$ becomes

$$B_1 \bigcirc B_2(H) = \frac{B_1(H) \times (1 - B_2(\overline{H}))}{1 - B_1(H) \times B_2(\overline{H})} \qquad (33)$$

It should be mentioned here that the sole purpose of the multiplication by $\dfrac{1}{1 - B_1(H) \times B_2(\overline{H})}$ is to obtain a *normalized* bpa. Therefore the multiplier will be referred to as the *normalization factor*.

We are now in a position to relate our formalism to these concepts. As mentioned earlier, we may identify our degree of belief $Bel(H|E)$ with Dempster-Shafer's belief function $B(H)$. (Equivalently $m(H)$, for that matter.) In order to emphasize that $B(H)$ has been obtained from the knowledge $E$, we shall rewrite it as $B_E(H)$. Similarly, $Bel(H|F$ is $V_H)$ can be written as $B_F(H)$, with the understanding that the factor $F$ exists to the whole by taking its value at $V_H$. Finally, our joint degree of belief, namely $Bel(H|E, F$ is $V_H)$, can be regarded as the combined belief, in the sense of Dempster's rule, from the two component beliefs $Bel(H|E)$ and $Bel(H|F$ is $V_H)$. Stated simply, we shall equate $Bel(H|E, F$ is $V_H)$ with $B_E \bigcirc B_F(H)$ in this section. Now that the two body of concepts have been interpreted as above, we are then ready to show that our combination method for multiple supportive factors is consistent with the Dempster-Shafer result. Moreover, we shall show that our combination of adverse factors, or that between supportive and adverse factors, corresponds to an *unnormalized* Dempster's formula.

In the first place, let us demonstrate that under an additional assumption the Dempster's equation in (32) can be used to derive our formula in (10) for combining two supportive factors. To this end, the assumption we have to make is that the degree of support ($SUPP$) by the factor $F$ in our equation (2) can be equated to the belief $B_F(H)$ in Dempster-Shafer's terms. As a result of this equality, with the interpretations we made earlier, the upper equations in (8) and (9) may be rewritten as

$$B_E \bigcirc B_{F_1}(H) = B_E(H) + B_{F_1}(H) \times (1 - B_E(H)) \qquad (34)$$

$$B_E \bigcirc B_{F_2}(H) = B_E(H) + B_{F_2}(H) \times (1 - B_E(H)) \qquad (35)$$

(which exactly re-formulate the Dempster's equation in (32)). The next step is to recognize that our joint belief $Bel(H|E, F$ is $V_{H_1}, F$ is $V_{H_2})$ is actually $B_E \bigcirc [B_{F_1} \bigcirc B_{F_2}](H)$. To compute it, we can make use of the *associativity* in the general Dempster's rule, (which can also been seen from (32) itself) and replace $B_E(H)$ in (35) by $B_E \bigcirc B_{F_1}(H)$ of (34). Recalling that this substitution process is exactly the way we have derived the upper equation in (10), we therefore have completed the proof.

Before turning to cases involving adverse factors, it is useful to remark that the normalization factor in (33) can lead to counter-intuitive results, as has been pointed out by [Zadeh 1979]. In fact, as mentioned earlier, our model corresponds to an unnormalized form of (33); that is, $B_1 \bigcirc B_2(H) = B_1(H) \times (1 - B_2(\overline{H}))$. To establish this correspondence, we have to make the following assumption that, the degree of adversity $(ADV)$ by the factor $F$ in our equation (4) equals the disbelief $B_F(\overline{H})$ in Dempster-Shafer's terms. This being the case, our combining formulae in (12) and (13) for adverse factors can be seen to become instances of an unnormalized equation (33). Then our joint posterior belief in two adverse factors, which has been equated to $B_E \bigcirc [B_{F_1} \bigcirc B_{F_2}](H)$, can be obtained by the same substitution process between (12) and (13) as in the case of supportive factors. (This is all because of the associativity of Demspter's rule.) As a result, the counterpart of our combining formula in (14) is derived:

$B_E \bigcirc [B_{F_1} \bigcirc B_{F_2}](H) = B_E(H) \times (1 - B_{F_1}(\overline{H})) \times (1 - B_{F_2}(\overline{H}))$, which is what we have claimed. By the same token, the case in which supportive factors are combined with adverse factors can also be established to validate our formula in (16).

To conclude this section, we note that, although a correspondence exists between Dempster-Shafer's result and our model on the supportive and the adverse factors, the correspondence is certainly not a comprehensive one. Beside the fact that some additional assumptions were made in the process of argument, it is apparent that the concept of factor margins and that of partial matching extends beyond their framework. In a nutshell, this is because what was addressed there is the issue of combination of beliefs in the realm of hypotheses, not the issue of relations between factors and decisions. (This explains in part the reasons for not adapting their theory in the beginning, let alone the rest of classes of roles.)

### 4. Concluding Remark

In review of what we have presented in the context of subjective decision making, we started with a role classification system for all the decision factors. Each role of the factor has two margin values and an intensity to cast its peculiar influence on the final degree of belief. Then combinations between factors were studied. Based on the evidence/factor dichotomy, we embodied the concept of partial matching in the belief system and went on to derive corresponding equations for evidential aggregation. Finally, we took certain interpretations and demonstrated the similarity to, and the distinction from Dempster-Shafer theory in special cases. Notwithstanding the apparant resemblance, it then became clear that the theory really addresses different issues than ours.

It should be noted that the entirety of our belief system may very well have been presented without change under the framework of *fuzzy-set-*theoretic *possibility* theories (e.g. [Zadeh 1965], [Zadeh 1978b]). To adapt the formalism to the new setting, we should regard $Bel(H|E)$ as the possibility that the decision maker, with all the evidence $E$, will grant a positive decision on the case being considered. Note that the concept of possibility is different from that of the probability. (It expresses the ease of attainment as a matter of degree.) This observation manefests the general applicability of the role system.

In practiccal use, however, we expect that difficulties may arise for the decision maker. Especially when asked to give subjective opinions in terms of *precise numbers* on how a factor (with its intensity and two margins) would influence the decision. Fortunately, this problem can be circumvented by employing linguistic descriptions [Zadeh 1975]. A number of empirical studies (e.g. [Freksa 1982]) have clearly spelled out the merits of such a linguistic, rather than numerical approach. Again, the extension to these fuzzy-set-based linguistic terms doesn't change the resulting formulae in our belief system.

To conclude, the rich semantics of the role system along with the computational belief mechanisms provide us with a deeper level of expression for factor causality. The versatile implications by kinds of evidence can now be drawn more naturally than using classical decision theories. Thus the integrity of a decision maker's cognitive structure may be better preserved, meriting the system a more realistic mentality.

### Acknowledgement

The author is indebted to Professor Zadeh for his continuous encouragement. Recent discussions with John Yen have greatly benefited the research. The author also thanks Dr. Chris Talbot and Dr. Malcolm Slaney at Schlumberger for their prompt help in preparing the paper.

### Reference

[Adams 1984] Adams, J.B., "Probabilistic Reasoning and Certainty Factors," *Rule-Based Expert Systems* (Buchanan, B. and Shortliffe, E., eds.). Menlo-Park: Addison-Wesley, pp. 263-271, 1984.

[Carnap 1950] Carnap, R., "The Two Concepts of Probability," *Logical Foundations of Probability*. Chicago: University of Chicago Press, pp. 19-51, 1950.

[Cohen 1983] Cohen, P., *Heuristic Reasoning about Uncertainty: An Artificial Intelligence Approach,* PhD dissertation, Dept. of Computer Science, Stanford University, 1983.

[Dempster 1967] Dempster, A.P., "Upper and Lower Probabilities Induced by a Multivalued Mapping," *Ann. Math. Statist.*, Vol. 38 (1967), pp. 325-329.

[Dombi 1982] Dombi, J., "A General Class of Fuzzy Operations," *Fuzzy Sets and Systems*, Vol. 8 (1982), pp. 149-163.

[Duda 1979] Duda, R.O., Hart, P.E., Konolige, K., and Reboh, R, *A Computer-Based Concultant for Mineral exploration*, SRI Report, Sept., 1979.

[Edwards 1954] Edwards, W., "The Theory of Decision Making," *Psychological Bulletin*, Vol. 51, No. 4 (1954), pp. 380-417.

[Fishburn 1970] Fishburn, P.C., *Utility Theory and Decision Making*. New York: John Wiley, 1970.

[Fishburn 1967] Fishburn, P.C., "Methods of Estimating Additive Utilities," *Management Sci.*, Vol. 13 (1967), pp. 435-453.

[Freksa 1982] Freksa, C., "Linguistic Description of Human Judgements in Expert Systems and in the 'Soft' Sciences," *Approximate Reasoning in Decision Analysis* (M. M. Gupta and E. Sanchez, eds.). New York: North-Holland, pp. 297-305, 1982.

[Ishizuka 1981] Ishizuka, M., Fu, K.S., and Yao, J.P.T., "A Theoretical Treatment of Certainty Factor in Production Systems," Memo CE-SIR-81-6, Purdue Univ., 1981.

[Gordon 1984] Gorden, J., and Shortliffe, E., "The Dempster-Shafer Theory of Evidence," *Rule-Based Expert Systems* (Buchanan, B. and Shortliffe, E., eds.). Menlo-Park: Addison-Wesley, pp. 272-292, 1984.

[Krantz 1971] Krantz, D.H., Luce R.D., and Tversky, A., *Foundations of Measurement*, Vol. 1. New York: Academic Press, 1971.

[Michalski 1980] Michalski, R.S., and Chilausky, R.L., "Learning by Being Told and Learning from Examples: An Experimental Comparison of the Two Methods of Knowledge Acquisition in the Context of Developing an Expert System for Soybean Disease Diagnosis," *Int. J. Policy Analysis and Information Systems*, Vol. 4, No. 2(June 1980), pp. 125-161.

[Prade 1985] Prade, H., "A Computational Approach to Approximate and Plausible Reasoning with Applications to Expert Systems," *IEEE Trans. Pattern Analysis and Machine Intelligence*, Vol. 7, No. 3(May, 1985), pp. 260-283.

[Simon 1947] Simon, H.A., *Administrative Behavior*. New York: Macmillan, 197.

[Shafer 1976] Shafer, G., *A Mathematical Theory of Evidence*. Princeton: Princeton University Press, 1976.

[Shortliffe 1976] Shortliffe, E.H., *Computer Based Medical Consultations: MYCIN*. New York: American Elsevier, 1976.

[Tversky 1972] Tversky, A., "Eliminatation by Aspects: A Theory of Choice," *Psychology Review* (1979), pp. 281-299.

[van Melle 1980] van Melle, W., *A Domain Independent System that Aids in Constructing Knowledge Based Consultation Programs*, PhD dissertation, Dept. of Computer Science, Stanford Univ., 1980.

[von Neumann 1947] van Neumann, J., and Morgenstern, O., *Theory of Games and Economic Behavior*. Princeton: Princeton University Press, 1947.

[Yager 1977] Yager, R.R., "Multiple Objective Decision-Making Using Fuzzy Sets," *Int. J. Man-Machine Studies*, Vol. 9 (1977), pp. 375-382.

[Yen 1985] Yen, John, *private communications*, Dept. of EECS, U.C. Berkeley, April, 1985.

[Zadeh 1965] Zadeh, L.A., "Fuzzy Sets," *Informat. and Control* (June 1965), pp. 338-353.

[Zadeh 1975] Zadeh, L.A., "The Concept of a Linguistic Variable and Its Application to Approximate Reasoning. Part I," *Information Sci.*, Vol. 8 (1975), pp. 199-249; Part II, *Information Sci.*, Vol. 8 (1975), pp301-357; Part III, *Information Sci.*, Vol. 9 (1975), pp43-80.

[Zadeh 1978a] Zadeh, L.A., "PRUF--a Meaning Representation Language for Natural Languages," *Int. J. Man-Machine Studies*, Vol. 10(1977), pp. 395-460.

[Zadeh 1978b] Zadeh, L.A., "Fuzzy Sets as a Basis for a Theory of Possibility," *Fuzzy Sets and Systems*, Vol. 1 (1978), pp. 395-460.

[Zadeh 1979] Zadeh, L.A., "On the Validity of Dempster's Rule of Combination of Evidence," Memo M79/24, Electronics Research Lab., U.C. Berkeley, 1979.